\documentclass[runningheads]{llncs}
\usepackage[utf8]{inputenc}
\usepackage[T1]{fontenc}
\usepackage[english,french]{babel}
\usepackage{graphicx}
\usepackage[group-separator={,}]{siunitx}
\begin{document}
\title{IAM at CLEF eHealth 2018: Concept
Annotation and Coding in French Death Certificates}
\author{Sébastien Cossin\inst{1,}\inst{2}%\orcidID{0000-0002-3845-8127}
, Vianney Jouhet\inst{1,}\inst{2}, Fleur Mougin\inst{1}, Gayo Diallo\inst{1},\\
Frantz Thiessard\inst{1,}\inst{2}
}

%\titlerunning{IAM at CLEF eHealth 2018}
%\authorrunning{Cossin S et al.}
\pagestyle{empty}

\institute{
Univ. Bordeaux, Inserm, Bordeaux Population Health Research Center, team ERIAS, UMR 1219, F-33000 Bordeaux, France \\
\texttt{sebastien.cossin@u-bordeaux.fr} 
  \and 
 CHU de Bordeaux, Pôle de santé publique, Service d'information médicale,  Informatique et Archivistique Médicales (IAM), F-33000 Bordeaux, France \\
%\texttt{Marc.Lun@ici.fr} 
% \texttt{Michele.Lautre@la.fr} 
}
\maketitle              % typeset the header of the contribution
\selectlanguage{english}

\begin{abstract}
In this paper, we describe the approach and results for our participation in the task 1 (multilingual information extraction) of the CLEF eHealth 2018 challenge. We addressed the task of automatically assigning ICD-10 codes to French death certificates. We used a dictionary-based approach using materials provided by the task organizers. The terms of the ICD-10 terminology were normalized, tokenized and stored in a tree data structure. The Levenshtein distance was used to detect typos. Frequent abbreviations were detected by manually creating a small set of them. Our system achieved an F-score of 0.786 (precision: 0.794, recall: 0.779). These scores were substantially higher than the average score of the systems that participated in the challenge. 

  \keywords{Semantic annotation \and Entity recognition \and Natural Language Processing \and Death certificates}
\end{abstract}

\selectlanguage{french}

\section{Introduction}
In this paper, we describe our approach and present the results for our participation in the task 1, \textit{i.e.} multilingual information extraction, of the CLEF eHealth 2018 challenge~\cite{CLEFeHealth}. More precisely, this task consists in automatically coding death certificates using the International Classification of Diseases, 10th revision (ICD-10)~\cite{task1}.

We addressed the challenge by matching ICD-10 terminology entries to text phrases in death certificates. Matching text phrases to medical concepts automatically is important to facilitate tasks such as search, classification or organization of biomedical textual contents~\cite{jovanovic}. 
Many concept recognition systems already exist~\cite{jovanovic,Noble}. They use different approaches and some of them are open source. 
We developed a general purpose biomedical semantic annotation tool for our own needs. The algorithm was initially implemented to detect drugs in a social media corpora as part of the Drugs-Safe project~\cite{Bigeard}. We adapted the algorithm for the ICD-10 coding task. The main motivation in participating in the challenge was to evaluate and compare our system with others on a shared task.

\section{Methods}

In the following subsections, we describe the corpora, the terminology used, the steps of pre-processing and the matching algorithm. 
\subsection{Corpora}
The data set for the coding of death certificates is called the CépiDC corpus.
Three CSV files (AlignedCauses) were provided by task organizers containing annotated death certificates for different periods : 2006 to 2012, 2013 and 2014. This training set contained \num{125383} death certificates. Each certificate contains one or more lines of text (medical causes that led to death) and some metadata.
Each CSV file contains a "Raw Text" column entered by a physician, a "Standard Text" column entered by a human coder that supports the selection of an ICD-10 code in the last column. Table~\ref{table:tableau} presents an excerpt of these files. Zero to multiples ICD-10 codes can be assigned to each line of a death certificate.

\begin{table}[ht]
\centering
\begin{tabular}{p{0.6\linewidth}cc}
\hline
  Raw Text & Standard Text & ICD-10 code \\ 
  \hline
  SYNDROME DE GLISEMENT AVEC GRABATISATION DEPUIS OCTOBRE 2012 & syndrome glissement & R453 \\ 
  \hline
  SYNDROME DE GLISEMENT AVEC GRABATISATION DEPUIS OCTOBRE 2012 & grabatisation 2 mois & R263 \\ 
   \hline
\end{tabular}
\caption{One raw text sample with three selected columns of the training data. Raw Text: text entered by a physician (duplicated in the file when multiple codes are assigned). 
\newline
Standard Text: text entered by a human coder to support the selection of the ICD-10 code}
 \label{table:tableau}
\end{table}

\subsection{Dictionaries}
We constructed two dictionaries based on ICD-10. In practice, we selected all the terms in the "Standard Text" column of the training set to build the first one which was used in the second run. In the first run, we added to this previous set of terms the 2015 ICD-10 dictionary provided by the task organizers. This dictionary contained terms that were not present in the training corpus. 
When a term was associated with multiple ICD-10 codes in our dictionary, we kept the most frequent one (Table~\ref{table:occurences}). 

The first dictionary contained \num{42439} terms and 3,539 ICD-10 codes (run2) and the second one \num{148448} terms and 6,392 ICD-10 codes (run1). 

Metadata on death causes were not used (age, gender, location of death). 

% latex table generated in R 3.3.2 by xtable 1.8-2 package
% Thu May 24 08:50:39 2018
\begin{table}[ht]
\centering
\begin{tabular}{llr}
  \hline
Standard Text & ICD-10 code & number of occurrences \\ 
  \hline
  avc & F179 &   1 \\ 
  avc & I64 & 260 \\ 
  avc & I640 & 1,635 \\ 
  avc & T821 &   1 \\ 
  avc & Z915 &   1 \\ 
  avc & I489 &   1 \\ 
   \hline
   \end{tabular}
   \caption{Some terms like "avc" were associated with multiple ICD-10 codes in our dictionary. We kept the most frequent ICD-10 code, I640 in this case. \newline
   Standard Text: text entered by a human coder to support the selection of an ICD-10 code. }
\label{table:occurences}
\end{table}

\subsection{Terms pre-processing}
\begin{figure}[ht!]
\begin{center}
\includegraphics[scale=0.70]{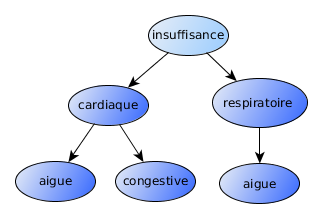}
\caption{Terms in our dictionaries are normalized, tokenized and stored in a tree data structure. Each dark blue node corresponds to a term. In this figure, five terms are described: "insuffisance cardiaque", "insuffisance cardiaque aigue", "insuffisance cardiaque congestive", "insuffisance respiratoire" and "insuffisance respiratoire aigue". The token "insuffisance" is the first token of many terms but it does not match any term by itself (light blue).}
\label{fig:radixtree}
\end{center}
\end{figure}

All the terms were normalized through accents (diacritical marks) and punctuation removal, lowercasing and stopwords removal (we created a list of 25 stopwords for this task). Then, each term was tokenized and stored in a tree data structure. Each token of a N-gram term is a node in the tree and N-grams correspond to different root-to-leaf paths~\cite{treeReference} (Figure~\ref{fig:radixtree}). 

\subsection{Matching algorithm}
\begin{figure}[ht!]
\begin{center}
\includegraphics[width=0.8\textwidth]{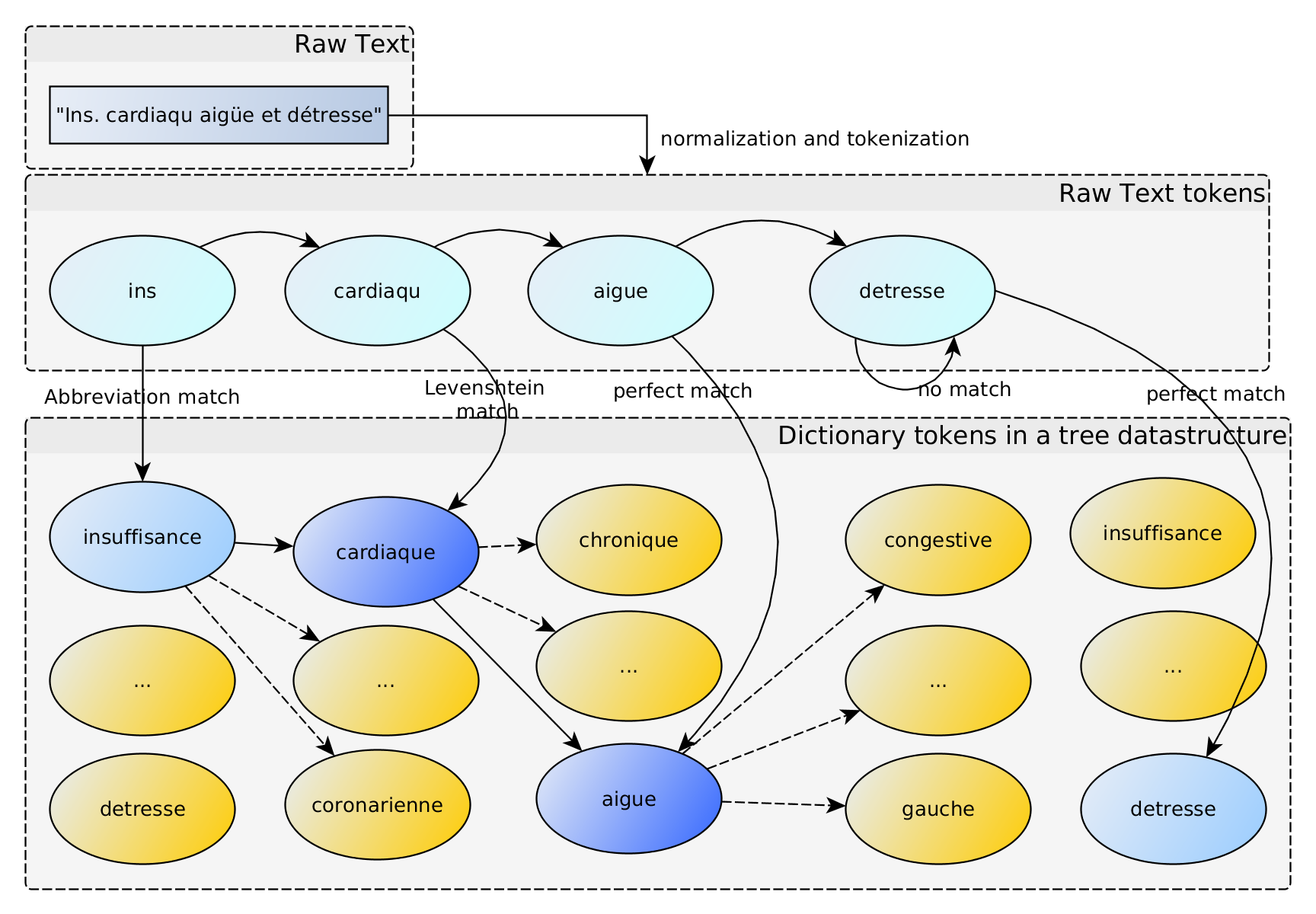}
\caption{Algorithm pipeline. The text was normalized and tokenized. For each token, the algorithm used three matching techniques to detect a token in the dictionary. Possible dictionary tokens depended on the current location in the tree. 
\newline Raw Text: a death cause entered by a physician. Solid arrows: path taken by the algorithm. Dotted arrows: other available tokens at this depth. Dark blue nodes : a term in the dictionary. Light blue nodes: token that does not match any term by itself}
\label{fig:algorithm}
\end{center}
\end{figure}

\begin{figure}[ht!]
\begin{center}
\includegraphics[width=0.7\textwidth]{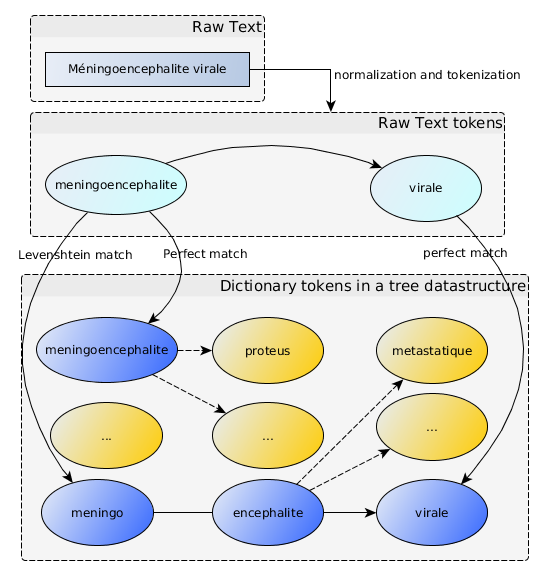}
\caption{The token "meningoencephalite" was matched to the unigram "meningoencephalite" by the perfect match method and to the bigram "meningo encephalite" by the Levenshtein method. The algorithm explored different paths in the tree. It detected the term "meningo encephalite virale" ("meningoencephalite virale" did not exist). Only the longest term was kept.}
\label{fig:algorithmbi}
\end{center}
\end{figure}

The goal of our algorithm was to recognize one or many dictionary entries in a raw text. An example is given in Figure~\ref{fig:algorithm}.
For each raw text entry, the same normalization steps described above were performed first. The raw text was then tokenized. For each token, the algorithm looked for an available dictionary token depending on where  it currently was in the tree. For example, the token "cardiaque" was possible after the token "insuffisance" but was not available at the root of the tree. 
\newline
For each token, the algorithm used three matching techniques: perfect match, abbreviation match and Levenshtein match. The abbreviation match technique used a dictionary of abbreviations. We manually added nine frequent abbreviations after looking at some examples. The Levenshtein matching technique used the Levenshtein distance. It corresponds to the minimum number of single-character edits (insertions, deletions or substitutions) required to change one token into the other. The Lucene\texttrademark implementation of the Levenshtein distance was used. 
\newline
In Figure~\ref{fig:algorithm}, the algorithm used these three techniques to match the tokens "ins", "cardiaqu", "aigue" to the dictionary term "insuffisance cardiaque aigue" whose ICD-10 code is I509. As the following token "detresse" was not a dictionary entry at this depth, the algorithm saved the previous and longest recognized term and restarted from the root of the tree. At this new level, "detresse" was detected but as no term was associated with this token alone, no ICD-10 code was saved. Finally, only one term was recognized in this example.
\newline
Besides unigrams, bigrams were also indexed in Lucene\texttrademark to resolve composed words. For example, "meningoencephalite" matched the dictionary entry "meningoencephalite" by a perfect match and "meningo encephalite" thanks to the Levensthein match (one deletion). Therefore, the algorithm entered two different paths in the tree (Figure~\ref{fig:algorithmbi}). By combining these different matching methods for each token, the algorithm was able to detect multiple lexical variants. 
The program was implemented in Java and the source code is on Github\footnote[1]{\url{https://github.com/scossin/IAMsystem}}. 

\section{Results}
We submitted two runs on the CépiDC test set, one used all the terms entered by human coders in the training set only (run 2), the other (run 1) added the 2015 ICD-10 dictionary provided by the task organizers to the set the terms of run 1. We obtained our best precision (0.794) and recall (0.779) with run 2. 

Table~\ref{tab:result} shows the performance of our system with median and average scores of all participants in this task.

\begin{table}[ht]
\centering
\begin{tabular}{lccc}
  \hline
System & Precision & Recall & F-measure \\ 
  \hline
  run1 & 0.782 & 0,772 & 0.777 \\ 
  run2 & 0.794 & 0.779 & 0.786 \\ 
  average score & 0.712 & 0.581 & 0.634 \\
  median score & 0.771 & 0.545 & 0.641 \\
   \hline
   \end{tabular}
   \caption{System performance on the CépiDC test set}
\label{tab:result}
\end{table}

\section{Discussion}

Surprisingly, adding more terms (run 1) did not improve the recall, which appears to be even slightly worse.
The results were quite promising for our first participation in this task, using a general purpose annotation tool. 

A limitation of the proposed algorithm that impacted recall was the absence of term detection when adjectives were isolated. For example, in the sentence "metastase hepatique et renale", "metastase renale" was not recognized even though the term existed. This situation seemed to be quite frequent.

Some frequent abbreviations were manually added to improve the recall in this corpora. Improvement at this stage may be possible by automating the abbreviation detection or by adding more entries manually.

In the past, other dictionary-based approaches performed better~\cite{neveolClinicalInformationExtraction2016a}. In 2016, the Erasmus system~\cite{Erasmus} achieved an F-score of 0.848 without spelling correction techniques. In 2017, the SIBM team~\cite{Cabot} used a dictionary-based approach with fuzzy matching methods and phonetic matching algorithm to obtain an F-score of 0.804. 

Further improvement may be possible by using a better curated terminology. We are currently investigating frequent irrelevant codes that may have impacted the precision. A post-processing filtering phase could improve the precision. 

We also plan to combine machine learning techniques with a dictionary-based approach. Our system can already detect and replace typos and abbreviations to help machine learning techniques increase their performance.

\section{Affiliation}

DRUGS-SAFE National Platform of Pharmacoepidemiology, France

\section{Funding}

The present study is part of the Drugs Systematized Assessment in real-liFe Environment (DRUGS-SAFE) research platform that is funded by the French Medicines Agency (Agence Nationale de Sécurité du Médicament et des Produits de Santé, ANSM). This platform aims at providing an integrated system allowing the concomitant monitoring of drug use and safety in France. The funder had no role in the design and conduct of the studies; collection, management, analysis, and interpretation of the data; preparation, review, or approval of the manuscript; and the decision to submit the manuscript for publication. This publication represents the views of the authors and does not necessarily represent the opinion of the French Medicines Agency.


\selectlanguage{english}

%\renewcommand{\refname}{References}
%\bibliographystyle{unsrt}

%\bibliography{IAS2018}

\begin{thebibliography}{8}
%\bibliographystyle{splncs04}

\bibitem{CLEFeHealth}
Suominen, H. and Kelly, L. and Goeuriot, L. and Kanoulas, E. and Azzopardi, L. and Spijker, R. and Li, D. and Névéol, A. and Ramadier, L. and Robert, A. and Palotti, J. and Jimmy and Zuccon, G.: Overview of the CLEF eHealth Evaluation Lab 2018. CLEF 2018 - 8th Conference and Labs of the Evaluation Forum, Lecture Notes in Computer Science (LNCS). Springer. (2018).

\bibitem{task1}
Névéol, A. and Robert, A. and Grippo, F. and Lavergne, T. and Morgand C. and Orsi, C. and Pelikán L. and Ramadier, L. and Rey, G. and Zweigenbaum,P: CLEF eHealth 2018 Multilingual Information Extraction task Overview: ICD10 Coding of Death Certificates in French, Hungarian and Italian. CLEF 2018 Evaluation Labs and Workshop: Online Working Notes, CEUR-WS, September, 2018.


\bibitem{jovanovic}
Jovanovi{\'c}, J. and Bagheri, E.: Semantic Annotation in Biomedicine: The Current Landscape. Journal of Biomedical Semantics (2017).
\doi{10.1186/s13326-017-0153-x}

\bibitem{Noble}
Tseytlin, E. and Mitchell, K. and Legowski, E. and Corrigan, J. and Chavan, G. and Jacobson, RS.: BMC bioinformatics (2016).
NOBLE - Flexible Concept Recognition for Large-Scale Biomedical Natural Language Processing. 
\doi{10.1186/s12859-015-0871-y}

\bibitem{Bigeard}
Bigeard, E.: Construction de lexiques pour l'extraction des mentions de maladies dans les forums de santé. TALN (2017).


\bibitem{treeReference}
Pibiri, GE. and Venturini, R.: Efficient Data Structures for Massive N-Gram Datasets. In: 40th International ACM SIGIR Conference on Research and Development in Information Retrieval, 
pp. 615--624. ACM, Shinjuku, Tokyo, Japan (2017)
\doi{10.1145/3077136.3080798}

\bibitem{neveolClinicalInformationExtraction2016a}
Névéol, A. and Cohen, KB. and Grouin, C. and Hamon, T. and Lavergne, T. and Kelly, L. and Goeuriot, L. and Rey, G. and Robert, A. and Tannier, X. and Zweigenbaum, P.:
Clinical Information Extraction at the CLEF eHealth Evaluation Lab 2016. CEUR workshop proceedings. (2016).


\bibitem{Erasmus}
Van Mulligen, E. and Afzal, Z. and Akhondi, S. and Vo, D. and Kors, J.: 
Erasmus MC at CLEF eHealth 2016: Concept Recognition and Coding in French Texts. Online Working Notes. CEUR-WS. (2016).


\bibitem{Cabot}
Cabot, C. and Soualmia, LF. and Darmoni, S.: 
SIBM at CLEF eHealth Evaluation Lab 2017: Multilingual Information Extraction with CIM-IND. CEUR-WS. (2017).

\end{thebibliography}

\end{document}